\documentclass{article} 
\usepackage{iclr2025_conference,times}

\usepackage{amsmath,amsfonts,bm}

\def\eqref#1{equation~\ref{#1}}

\def\1{\bm{1}}

\DeclareMathAlphabet{\mathsfit}{\encodingdefault}{\sfdefault}{m}{sl}
\SetMathAlphabet{\mathsfit}{bold}{\encodingdefault}{\sfdefault}{bx}{n}

\usepackage{hyperref}
\usepackage{url}

\usepackage{lineno}
\usepackage{amsmath}
\usepackage{array}    
\usepackage{makecell} 
\usepackage{ragged2e}
\usepackage{graphicx}
\usepackage{multirow}
\usepackage{booktabs}
\usepackage{tabularx, booktabs}

\title{Exploring the Effect of Reinforcement Learning on Video Understanding: Insights from SEED-Bench-R1
}

\author{
\resizebox{\textwidth}{!}{\textbf{Yi Chen}$^{1,2}$, \textbf{Yuying Ge}$^{2\dag}$, \textbf{Rui Wang}$^{3}$, \textbf{Yixiao Ge}$^{2\dag}$, \textbf{Lu Qiu}$^{1,2}$, \textbf{Ying Shan}$^2$, \textbf{Xihui Liu}$^{1\dag}$}\\
    $^1$The University of Hong Kong,
    $^2$ARC Lab, Tencent PCG, 
    $^3$The Chinese University of Hong Kong\\
\\
\centerline{\href{https://github.com/TencentARC/SEED-Bench-R1}{https://github.com/TencentARC/SEED-Bench-R1}}
}

\footnotetext{Corresponding Authors.}

\iclrfinalcopy 
\begin{document}

\maketitle

\vspace{-25pt}
\begin{figure}[!h]
    \centering
    \includegraphics[width=\textwidth]{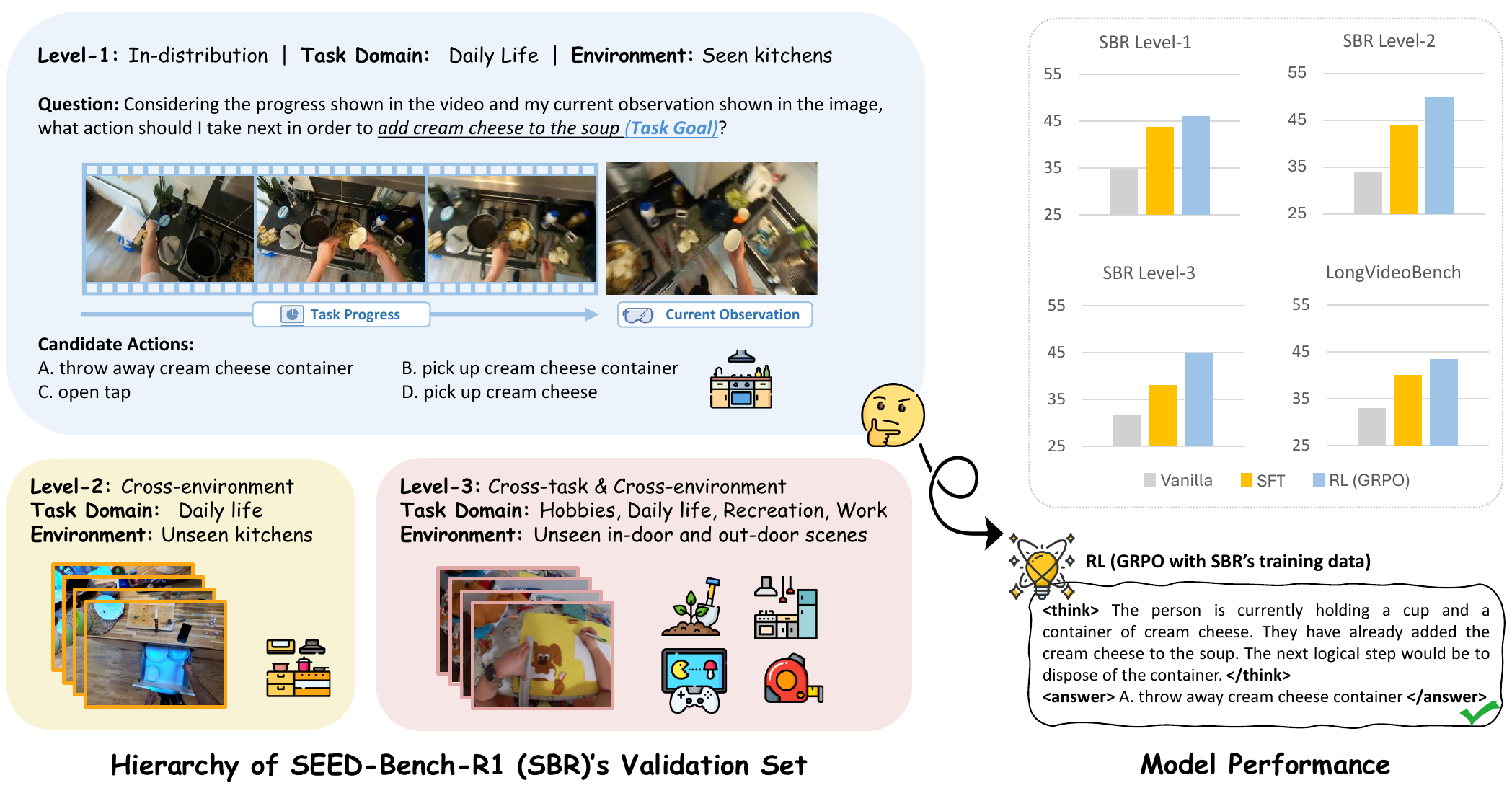}
    \vspace{-15pt}
    \caption{
    (a) Overview of SEED-Bench-R1 (SBR), which systematically evaluates post-training methods for MLLMs in video understanding. SBR features a three-level evaluation hierarchy: in-distribution, cross-environment, and cross-environment-task scenarios, equipped with training data containing easily verifiable ground-truth answers, to \textit{assess generalization across different levels}. These tasks necessitate both \textbf{perception and reasoning} to tackle complex real-world challenges. (b) Notably, an MLLM trained using reinforcement learning via GRPO outperforms both the base model and supervised fine-tuning (SFT) model, particularly in out-of-distribution scenarios (e.g., levels 2–3). Additionally, this RL-trained model exhibits strong generalization capabilities across general video understanding benchmarks (e.g., LongVideoBench).}
    \label{fig:teaser}
\end{figure}

\begin{abstract}
Recent advancements in Chain of Thought (COT) generation have significantly improved the reasoning capabilities of Large Language Models (LLMs), with reinforcement learning (RL) emerging as an effective post-training approach. Multimodal Large Language Models (MLLMs) inherit this reasoning potential but remain underexplored in tasks requiring both perception and logical reasoning. To address this, we introduce SEED-Bench-R1, a benchmark designed to systematically evaluate post-training methods for MLLMs in video understanding. It includes intricate real-world videos and complex everyday planning tasks in the format of multiple-choice questions, requiring sophisticated perception and reasoning. SEED-Bench-R1 assesses generalization through a three-level hierarchy: in-distribution, cross-environment, and cross-environment-task scenarios, equipped with a large-scale training dataset with easily verifiable ground-truth answers. Using Qwen2-VL-Instruct-7B as a base model, we compare RL with supervised fine-tuning (SFT), demonstrating RL's data efficiency and superior performance on both in-distribution and out-of-distribution tasks, even outperforming SFT on general video understanding benchmarks like LongVideoBench. Our detailed analysis reveals that RL enhances visual perception but often produces less logically coherent reasoning chains. We identify key limitations such as inconsistent reasoning and overlooked visual cues, and suggest future improvements in base model reasoning, reward modeling, and RL robustness against noisy signals. 
\end{abstract}

\section{Introduction}
\label{sec:intro}
Recent advancements in the reasoning capabilities of Large Language Models (LLMs)~\citep{guo2025deepseek, openai_o1, team2025kimi} have been driven by progress in long Chain of Thought (COT) generation. Among the various approaches to enhancing long COT, reinforcement learning (RL)~\citep{shao2024deepseekmath, ouyang2022training, yeo2025demystifying} has emerged as a particularly effective post-training method. RL enables LLMs to refine their COT through self-evolution with verifiable rewards, resulting in models that excel at solving complex problems and demonstrate strong generalization in out-of-distribution (OOD) tasks.

Multimodal Large Language Models (MLLMs), which are built upon LLMs, inherit their reasoning potential while incorporating additional modules for processing multimodal inputs. 
This has led to a growing interest in whether RL can similarly enhance multimodal understanding in MLLMs~\citep{zhang2025r1,liu2025visual,meng2025mm}. However, existing research primarily focuses on image-based tasks, emphasizing either perception (e.g., detection and grounding) or logical inference (e.g., multimodal math problem solving). We argue that \textit{an ideal testbed for studying the impact of post-training methods on multimodal understanding should balance both perception and logical inference}, ensuring models can integrate these capabilities to derive accurate answers. Additionally, the generalization capability of MLLMs must be rigorously evaluated to assess the robustness of post-training methods.

To address this, we introduce \textbf{SEED-Bench-R1}, a benchmark designed for systematic evaluation of the effectiveness of post-training methods on video understanding.
As shown in Table~\ref{tab:benchmark_statistics}, SEED-Bench-R1 is built on our prior works, reusing the training and validation data from our EgoPlan-Bench~\citep{chen2023egoplan}, as well as the test data from our EgoPlan-Bench2~\citep{chen2023egoplan}. Generally, the benchmark leverages realistic egocentric videos~\citep{Damen2022RESCALING,grauman2022ego4d} capturing everyday human activities, requiring models to understand open-form task goals, track long-horizon visual progress, perceive complex environmental observations, and reason about next actions using world knowledge.
Specifically, the validation data from EgoPlan-Bench are used for Level-1 (in-distribution) and Level-2 (OOD, cross-environment) evaluation, while the test data from EgoPlan-Bench2 cover more general domains and are used for Level-3 (OOD, cross-environment-task) evaluation.

As shown in Figure~\ref{fig:teaser}, SEED-Bench-R1 features: 1) intricate real-world visual inputs necessitating \textbf{nuanced perception}, 2) diverse questions requiring \textbf{logical inference with common sense}, 3) rigorously partitioned validation sets to assess in-distribution (L1) and OOD (L2/L3) \textbf{generalization capabilities}, 
and 4) large-scale, automatically constructed training questions with \textbf{easily verifiable ground-truth answers} (the ground-truth answer comes from the actual next action occurring right after the current
observation in the original uncropped video).

We evaluate representative post-training methods on SEED-Bench-R1 using Qwen2-VL-Instruct-7B~\citep{wang2024qwen2} as the base model, comparing RL (specifically GRPO~\citep{shao2024deepseekmath}) with supervised fine-tuning (SFT). Our experiments reveal that RL is highly data-efficient, significantly improving performance on both in-distribution (L1) and OOD (L2/L3) questions, even with simple outcome-based rewards. RL's superiority over SFT is particularly pronounced in OOD scenarios and extends to general video understanding benchmarks like LongVideoBench~\citep{wu2024longvideobench}.

Our analysis further explores how RL influences COT generation and its impact on visual perception and logical inference. We find that RL enhances the model's ability to attend to visual content more effectively, particularly in OOD scenarios.
RL teaches the model to dynamically query visual inputs with COT tokens rather than memorizing superficial reasoning patterns, thus achieving better generalization performance. However, limitations remain, such as the model's occasional neglect of key visual cues due to limited perception granularity and the lack of logical coherence in generated reasoning chains without process supervision, which hinder transparency and performance.

\begin{table}[!t]
\resizebox{\textwidth}{!}{%
\small
\centering
\begin{tabular}{ccccccc}
\toprule
\textbf{Split} & \textbf{\# Samples} & \textbf{Domain} & \textbf{Cross-Environment} & \textbf{Cross-Task} & \textbf{Video Source} & \textbf{Benchmark Source} \\ 
 \midrule
Train & 50,269 & Daily life & - & - & Epic-Kitchens & EgoPlan-Bench \\ 
Val-L1 & 2,432 & Daily life & $\times$ & $\times$ & Epic-Kitchens & EgoPlan-Bench \\ 
Val-L2 & 923 & Daily life & $\surd$ & $\times$ & Ego4D & EgoPlan-Bench \\ 
Val-L3 & 1,321 & Hobbies, Daily life, Recreation, Work & $\surd$ & $\surd$ & Ego4D & EgoPlan-Bench2 \\
\bottomrule
\end{tabular}}
\caption{Data statistics of SEED-Bench-R1, which consists of a training set and a hierarchical three-level validation set for in-distribution, cross-environment, and cross-environment-task evaluations. }
\label{tab:benchmark_statistics}
\end{table}

\begin{figure}[!t]
    \centering
    \includegraphics[width=\textwidth]{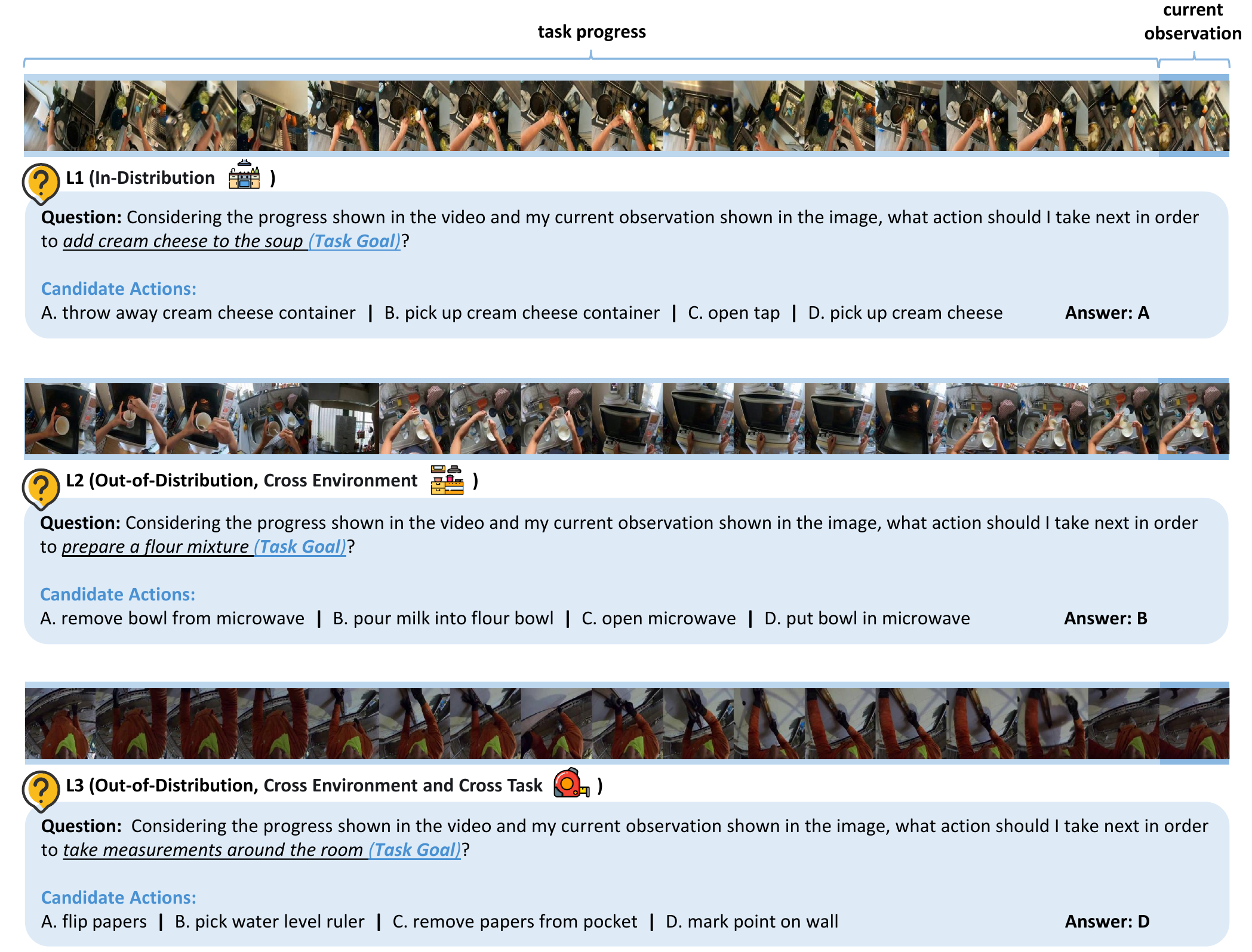}
    \caption{Example questions from the three-level evaluation hierarchy in SEED-Bench-R1’s validation set, including in-distribution, cross-environment, and cross-environment-task scenarios.}
    \label{fig:benchmark_overview}
    \vspace{-5pt}
\end{figure}

We conclude by outlining promising directions for future work, including eliciting stronger base reasoning abilities before RL, improving reward modeling to balance visual perception and logical inference, and enhancing RL's robustness to noisy reward signals. These steps are crucial for scaling RL to larger datasets and achieving more reliable alignment in multimodal understanding.

In summary, our contributions are as below:
\begin{itemize}

    \item We introduce SEED-Bench-R1, a video understanding benchmark designed to evaluate post-training methods, featuring real-world egocentric videos, questions balancing perception and logical inference, and a large-scale training dataset with rigorous validation splits divided into a three-level hierarchy for generalization assessment.

    \item Using SEED-Bench-R1, we conduct a systematic study and demonstrate that reinforcement learning (RL)—specifically GRPO—outperforms supervised fine-tuning (SFT) in data efficiency and generalization, notably in OOD scenarios, even with simple outcome-based rewards.

    \item Our analysis reveals that RL enhances video understanding by improving visual attention and encouraging dynamic querying of visual inputs via Chain of Thought (COT), though challenges remain in perception granularity and logical coherence.

    \item We identify future directions, such as strengthening the base model's reasoning capability pre-RL, refining reward design, and improving robustness to noise, to advance RL for multimodal alignment.
    
\end{itemize}

\section{Related Work}
Recent advancements in Large Language Models (LLMs), such as OpenAI's o1~\citep{openai_o1} and DeepSeek's R1~\citep{guo2025deepseek}, have demonstrated that reinforcement learning (RL) can significantly enhance reasoning abilities by enabling models to autonomously refine their Chain-of-Thought (CoT) processes without explicit supervision. Inspired by these successes, researchers have begun adapting RL-based approaches to Multimodal Large Language Models (MLLMs), achieving notable improvements in vision tasks~\citep{huang2025vision,zhao2025r1,liu2025visual,zhang2025r1}. However, existing work primarily focuses on image-based tasks—ranging from perception-heavy problems like image classification~\citep{lin2014microsoft} and object detection~\citep{gupta2019lvis} to reasoning-intensive tasks like visual math problem-solving~\citep{sun2024mm, lumathvista, zhang2024mathverse, zoudynamath}—while video understanding remains underexplored.

Preliminary efforts~\citep{wang-2025-open-r1-video,zhao2025r1} have applied RL to MLLMs for video benchmarks~\citep{wu2024longvideobench,liu2022mafw, jiang2020dfew}, but these evaluations often rely on narrow task formulations (e.g., recognizing a video’s dominant emotion~\citep{liu2022mafw, jiang2020dfew}) or suffer from limited training data~\citep{wu2024longvideobench}, hindering systematic analysis. While additional benchmarks exist to assess MLLMs across various dimensions~\citep{li2024mvbench,liu2024tempcompass,fang2024mmbench,fu2024video,song2024moviechat,chandrasegaran2024hourvideo,zhou2024mlvu,chen2024rextime,xiao2024can,cheng2025v}, none fully address the critical requirements for advancing video understanding.

Specifically, there is currently no comprehensive benchmark that fulfills three key criteria:
(1) \textbf{Large-scale training resources} to support robust post-training,
(2) \textbf{Structured validation sets} to assess different levels of generalization—essential for rigorously evaluating post-training methods like RL,
(3) \textbf{Diverse tasks} that balance perception and reasoning under real-world conditions aligned with human beings. 
To bridge this gap, we introduce \textbf{SEED-Bench-R1}, a video understanding benchmark designed to require balanced visual perception and logical reasoning across diverse real-world scenarios. Unlike prior work, our benchmark includes a large-scale training dataset and a carefully partitioned validation set with three generalization tiers, enabling comprehensive evaluation of MLLM capabilities and training methodologies.

\section{SEED-Bench-R1}

As shown in Figure~\ref{fig:teaser}, SEED-Bench-R1 is a benchmark designed for systematically studying the impact of post-training methods for MLLMs on video understanding. 
Built upon our previous works, EgoPlan-Bench~\citep{chen2023egoplan} and EgoPlan-Bench2~\citep{qiu2024egoplan}, SEED-Bench-R1 features 1) intricated visual inputs from the real world, 2) diverse questions requiring logical inference with common sense to solve practical tasks, 3) rigorous partition of validation sets to assess the robustness and generalization abilities of MLLMs across different levels, and 4) large-scale automatically constructed training questions with easily verifiable ground-truth answers.

As shown in Figure~\ref{fig:benchmark_overview}, the visual inputs and questions of SEED-Bench-R1 are grounded from realistic egocentric videos~\citep{Damen2022RESCALING,grauman2022ego4d} capturing everyday human activities. To answer questions in SEED-Bench-R1 correctly, the model must be capable of understanding open-form task goals, tracking long-horizon task progress, perceiving real-time environment state from an egocentric view, and utilizing inherent world knowledge to reason about the next action plan. The ground-truth answer comes from the actual next action occurring right after the current observation in the original uncropped video, with the negative options sampled from the same video. This challenging setting of candidate options demands a deep understanding of the environment state from dynamic visual input and world knowledge, such as action order dependency, rather than just the semantic meanings of task goals and actions, to discern the correct action plan. Moreover, the derivation of golden answers is traceable and easy to verify.

As listed in Table~\ref{tab:benchmark_statistics}, we provide both training and validation datasets to benefit community research. The training dataset is automatically constructed using Epic-Kitchens~\citep{Damen2022RESCALING} videos recording daily life household tasks in kitchen environments. The validation dataset has undergone strict human verification to ensure correctness and is divided into three levels. The Level-1 (L1) questions are created using the same video source as the training data, representing in-distribution evaluation scenarios where the visual environments and task goals share overlaps with the training data. The Level-2 (L2) questions cover similar task goals as L1, but the visual observations are recorded in unseen kitchen environments by new participators from the Ego4D~\citep{grauman2022ego4d} team. 
The Level-3 (L3) validation subset utilizes the full set of Ego4D videos beyond the kitchen-specific subset. It 
contains general-domain questions spanning hobbies, recreation, and work, in addition to daily life. The visual inputs come from a variety of in-door and out-door environments, posing greater challenges for testing the models' generalization abilities.

\section{Method}

We begin with Qwen2-VL-Instruct-7B~\citep{wang2024qwen2} as the base model to investigate whether reinforcement learning (RL) can effectively enhance the video understanding performance of Multimodal Large Language Models (MLLMs) on SEED-Bench-R1. For this exploration, we adopt GRPO~\citep{shao2024deepseekmath} as a representative RL algorithm and compare it with traditional post-training methods such as supervised fine-tuning (SFT). Both RL and SFT utilize 6k out of the 50k training samples from SEED-Bench-R1 for a preliminary study. To improve training efficiency, we limit the maximum number of sampled frames per input video to 16, with a frame resolution of $252 \times 252$. Besides, we explicitly append the frame indicating the current observation as an additional image input.

In the case of SFT, the training data is augmented with chain-of-thought (COT) reasoning processes for supervision. These COT annotations are distilled from stronger models, specifically Qwen2.5-VL-Instruct-7B and 72B, using rejection sampling. In contrast, RL eliminates the need for explicit COT annotations and instead relies on rule-based outcome supervision rewards. Similar to DeepSeek-R1~\citep{guo2025deepseek}, we employ a prompt template as following:

\begin{quote}
    ``\texttt{\{Question Here\}} Output the thinking process in \texttt{<think> </think>} and final answer in \texttt{<answer> </answer>} tags, i.e., \texttt{<think> reasoning process here </think><answer> answer here </answer>}''
\end{quote}

to guide the model to generate COT content before providing the final answer. This structured output facilitates easier extraction of answers for evaluation. 

\subsection{Supervised Fine-tuning (SFT)}

SFT is a simple post-training method, which refines LLMs by aligning their outputs with desired behaviors using human-curated data. The objective of SFT is to optimize the following loss function:

\begin{equation}
\small
\mathcal{J}_{S F T}(\theta)=\mathbb{E}\left[q, o \sim P_{s f t}(Q, O)\right]\left(\frac{1}{|o|} \sum_{t=1}^{|o|} \log \pi_{\theta}\left(o_{t} \mid q, o_{<t}\right)\right) \nonumber
\end{equation}

Here, the expectation is taken over input-output pairs \((q, o)\) sampled from a supervised dataset \(P_{sft}(Q, O)\). The goal is to maximize the average log-likelihood of the model generating the correct token \(o_t\) at each position \(t\), conditioned on the input \(q\) and the preceding tokens \(o_{<t}\). This process fine-tunes the model parameters \(\theta\) to minimize the discrepancy between the model's predictions and the ground truth, thereby improving task-specific performance.

\subsection{Outcome Supervision RL with Group Relative Policy Optimization (GRPO)}

GRPO is an RL framework initially developed to enhance the mathematical reasoning capabilities of LLMs. Unlike traditional PPO~\citep{ouyang2022training}, GRPO optimizes memory usage by eliminating the need for additional value function approximation. Instead, it simplifies the process by sampling multiple candidate responses for the same question and assessing their relative quality based on verifiable rewards. 

Specifically, for a given question \( q \) sampled from \(P_{sft}(Q)\), GRPO generates a group of \( G \) distinct responses \(\{o_1, o_2, ..., o_G\}\) using the policy model \( \pi_{\theta_{\text{old}}} \), rather than relying on predefined responses from \(P_{sft}(Q, O)\). The policy is then optimized by maximizing the following objective:
\setlength{\abovedisplayskip}{5pt}
\setlength{\belowdisplayskip}{5pt}
{\small
\begin{align*}
& \mathcal{J}_{G R P O}(\theta) =\mathbb{E}{\left[q \sim P_{sft}(Q),\left\{o_{i}\right\}_{i=1}^{G} \sim \pi_{\theta_{o l d}}(O \mid q)\right]}
\frac{1}{G} \sum_{i=1}^{G} \frac{1}{\left|o_{i}\right|} \sum_{t=1}^{\left|o_{i}\right|} \\
& \left\{\min \left[\frac{\pi_{\theta}\left(o_{i, t} \mid q, o_{i,<t}\right)}{\pi_{\theta_{o l d}}\left(o_{i, t} \mid q, o_{i,<t}\right)} \hat{A}_{i, t}, \operatorname{clip}\left(\frac{\pi_{\theta}\left(o_{i, t} \mid q, o_{i,<t}\right)}{\pi_{\theta_{o l d}}\left(o_{i, t} \mid q, o_{i,<t}\right)}, 1-\varepsilon, 1+\varepsilon\right) \hat{A}_{i, t}\right]-\beta \mathbb{D}_{K L}\left[\pi_{\theta}| | \pi_{r e f}\right]\right\}
\end{align*}}

Here, \( \epsilon \) and \( \beta \) are hyperparameters, \( D_{\text{KL}}(\pi_\theta || \pi_{\text{ref}}) \) represents the KL divergence between the trained policy \( \pi_\theta \) and the reference policy \( \pi_{\text{ref}} \), and \( \hat{A}_{i,t} \) is the per-token advantage calculated using outcome supervision based on relative rewards within the group. 
Specifically, for each response \( o_i \) to a question \( q \), a reward \( r_i \) is assigned based on rules by checking whether the answer extracted from \( o_i \) matches the ground-truth answer (e.g., \( r_i = 1 \) if correct, otherwise \( 0 \)). The rewards are normalized by computing their mean and standard deviation. Outcome supervision directly sets the advantages \( \hat{A}_{i, t} \) of all tokens in the response \( o_i \) to the normalized reward:

\begin{equation}
\small
\hat{A}_{i, t}=\widetilde{r}_{i}=\frac{r_{i}-\operatorname{mean}(\{r_1, ..., r_G\})}{\operatorname{std}(\{r_1, ..., r_G\})}. \nonumber
\end{equation}

\section{Experiments}

\subsection{Overall Performance}  
Table~\ref{tab:seedbench_r1_performance} summarizes the performance of the MLLM trained using different methods on SEED-Bench-R1. Notably, RL demonstrates remarkable data efficiency in improving the MLLM's performance on both in-distribution (L1) and OOD (L2 and L3) questions, despite relying solely on a simple outcome-based reward signal without specialized COT annotations. RL's superiority over SFT is particularly promising in OOD scenarios and even extends to a distinct video understanding benchmark (i.e., LongVideoBench~\citep{wu2024longvideobench} as shown in Table~\ref{tab:longvideobench_performance}).

\begin{table}[h]
\small
\centering
\resizebox{\textwidth}{!}{%
\begin{tabular}{lccccccc}
\toprule
\multirow{2}{*}{\textbf{Model}} & {\textbf{L1 (In-Distribution)}} & {\textbf{L2 (OOD, Cross-Environment)}} & \multicolumn{5}{c}{\textbf{L3 (OOD, Cross-Task, Cross-Environment)}} \\
\cmidrule(r){2-2} \cmidrule(r){3-3} \cmidrule(r){4-8}
 & \textbf{Daily Life} & \textbf{Daily Life} & \textbf{Daily Life} & \textbf{Hobbies} & \textbf{Recreation} & \textbf{Work} & \textbf{Overall} \\
\midrule
Vanilla & 34.79 & 34.02 & 31.21 & 32.30 & 33.33 & 30.69 & 31.57 \\
SFT & 43.79 & 44.10 & 38.27 & 41.02 & 32.24 & 38.61 & 38.15 \\
RL (GRPO) & \textbf{46.01} & \textbf{50.16} & \textbf{48.52} & \textbf{45.08} & \textbf{43.72} & \textbf{41.34} & \textbf{44.89} \\
\bottomrule
\end{tabular}}
\caption{Performance comparison on SEED-Bench-R1's hierarchical validation set. The model trained with GRPO using SEED-Bench-R1’s training set demonstrates significant performance improvements across both in-distribution (L1) and out-of-distribution (L2 and L3) scenarios.}
\label{tab:seedbench_r1_performance}
\end{table}

\begin{table}[h]
\small
\centering
\resizebox{\textwidth}{!}{%
\begin{tabular}{lccccccc}
\toprule
\multirow{2}{*}{\textbf{Model}} & \multicolumn{4}{c}{\textbf{Duration Group (unit: second)}} & \multicolumn{2}{c}{\textbf{Question Category}} & \multirow{2}{*}{\textbf{Overall}} \\ 
\cmidrule(r){2-5}  \cmidrule(rl){6-7} 
 & (8,15] & (15,60] & (180,600] & (900,3600] & L1 (Perception) & L2 (Relation) &  \\ 
 \midrule
Vanilla & 42.33 & 44.77 & 33.98 & 25.35 & 34.24 & 31.74 & 32.90 \\ 
SFT & 44.97 & \textbf{54.65} & 36.65 & 36.35 & 44.80 & 35.81 & 40.00 \\ 
RL (GRPO) & \textbf{54.50} & {53.49} & \textbf{42.48} & \textbf{37.23} & \textbf{48.48} & \textbf{38.90} & \textbf{43.40} \\  
\bottomrule
\end{tabular}
}
\caption{Performance comparison on LongVideoBench's validation set. The model trained with GRPO using SEED-Bench-R1's training set achieves higher performance gains on this general video understanding benchmark, which includes diverse themes.}
\label{tab:longvideobench_performance}
\end{table}

\begin{figure}[!t]
    \centering
    \begin{minipage}[b]{0.48\textwidth}
        \includegraphics[width=\textwidth]{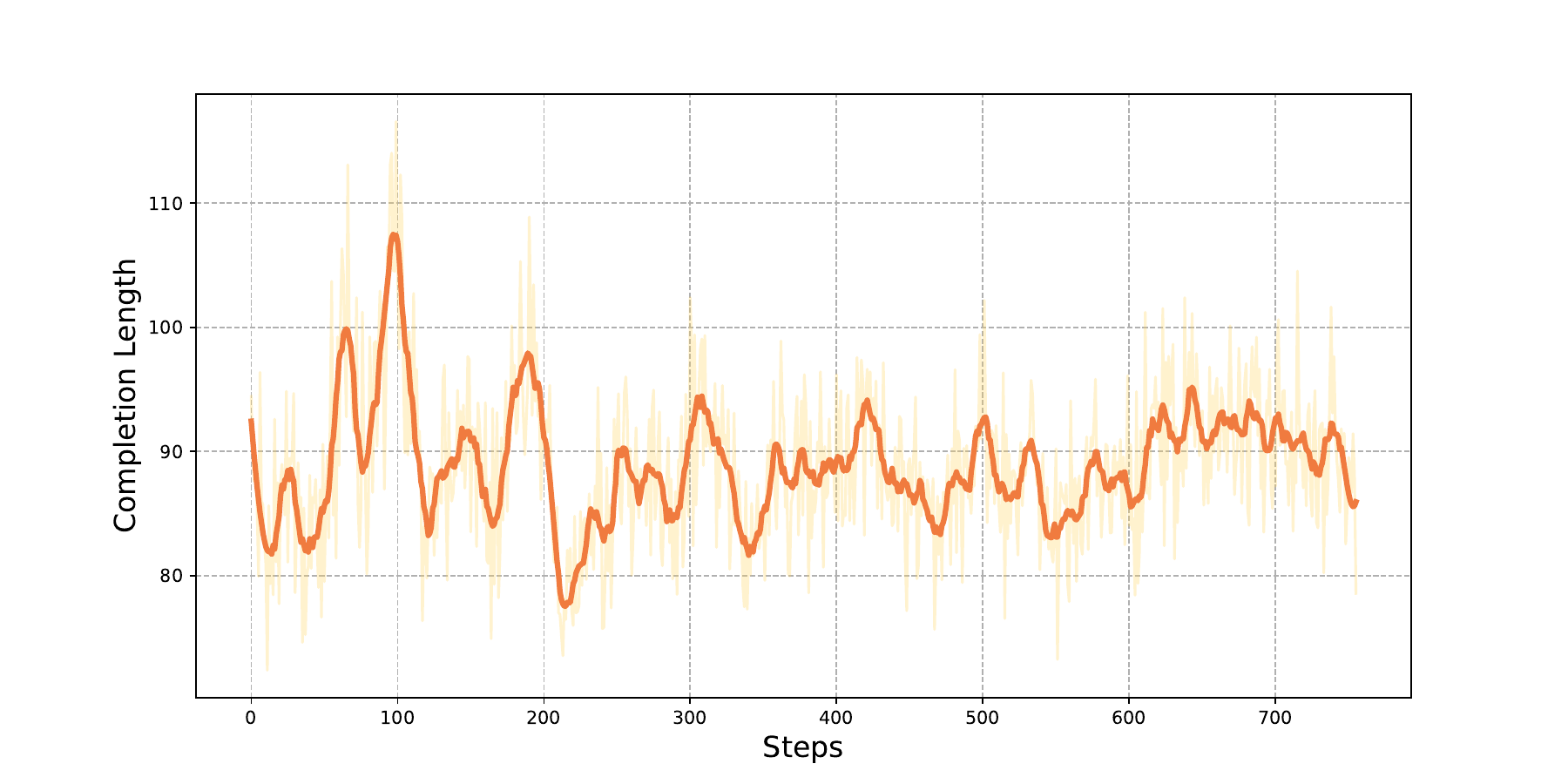}
    \end{minipage}
    \hfill
    \begin{minipage}[b]{0.48\textwidth}
        \includegraphics[width=\textwidth]{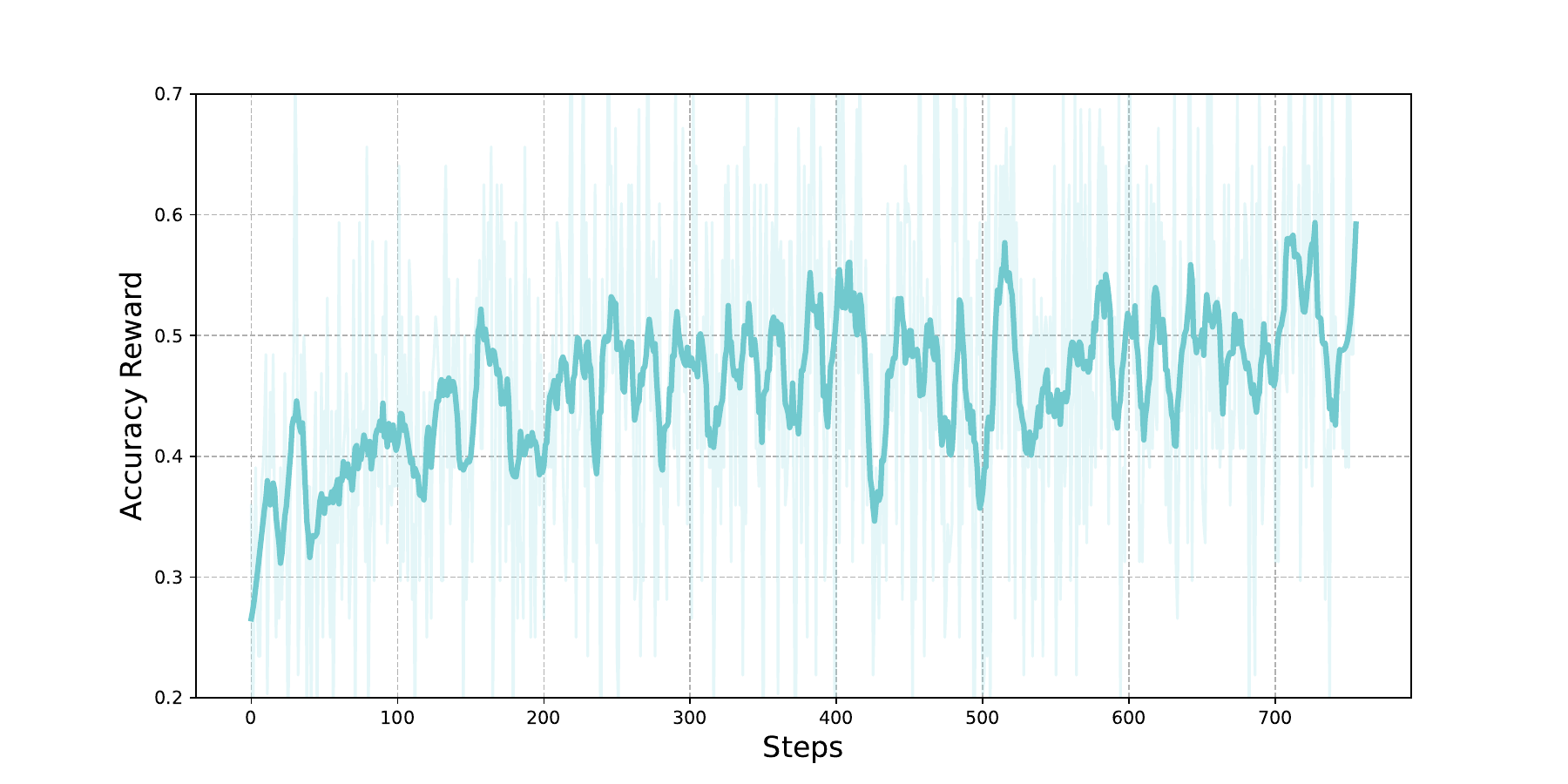}
    \end{minipage}
     \vspace{-5pt}
    \caption{
    The variation curves of completion length and accuracy reward w.r.t. RL training steps. While the reward value generally increases during RL, the completion length of the MLLM does not show a significant increase.
    }
    \label{fig:reward_completion_length}
\end{figure}

\begin{figure}[!t]
    \centering
    \includegraphics[width=\textwidth]{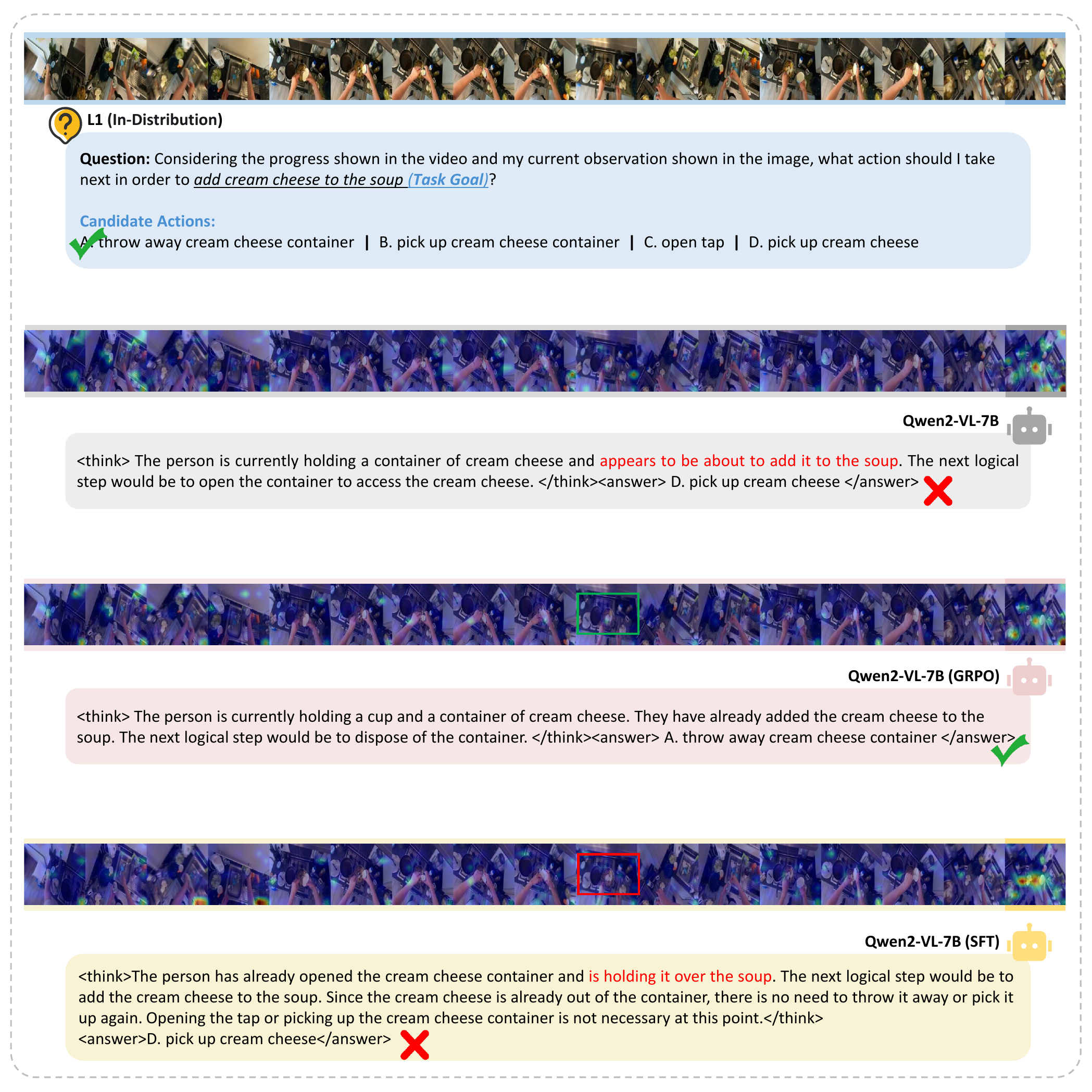}
    \caption{
    Comparison of model responses to a Level-1 question from SEED-Bench-R1. The visual input includes 16 sampled frames from a video (showing task progress) and a final observation image. Attention maps (output-to-visual tokens) are shown for each model: the base (Qwen2-VL-7B), GRPO fine-tuned, and SFT fine-tuned versions. The base and SFT models \textbf{exhibited illusory perceptions} (red text), while the GRPO model \textbf{attended more accurately to visual regions}—e.g., correctly identifying cream cheese in the pot (green box) and suggesting the next step (discarding the empty yogurt container). The SFT model’s attention was ineffective (red box), and the base model’s attention was dispersed, impairing judgment.
    }
    \vspace{-15pt}
    \label{fig:case_study_L1}
\end{figure}

\begin{figure}[!t]
    \centering
    \includegraphics[width=\textwidth]{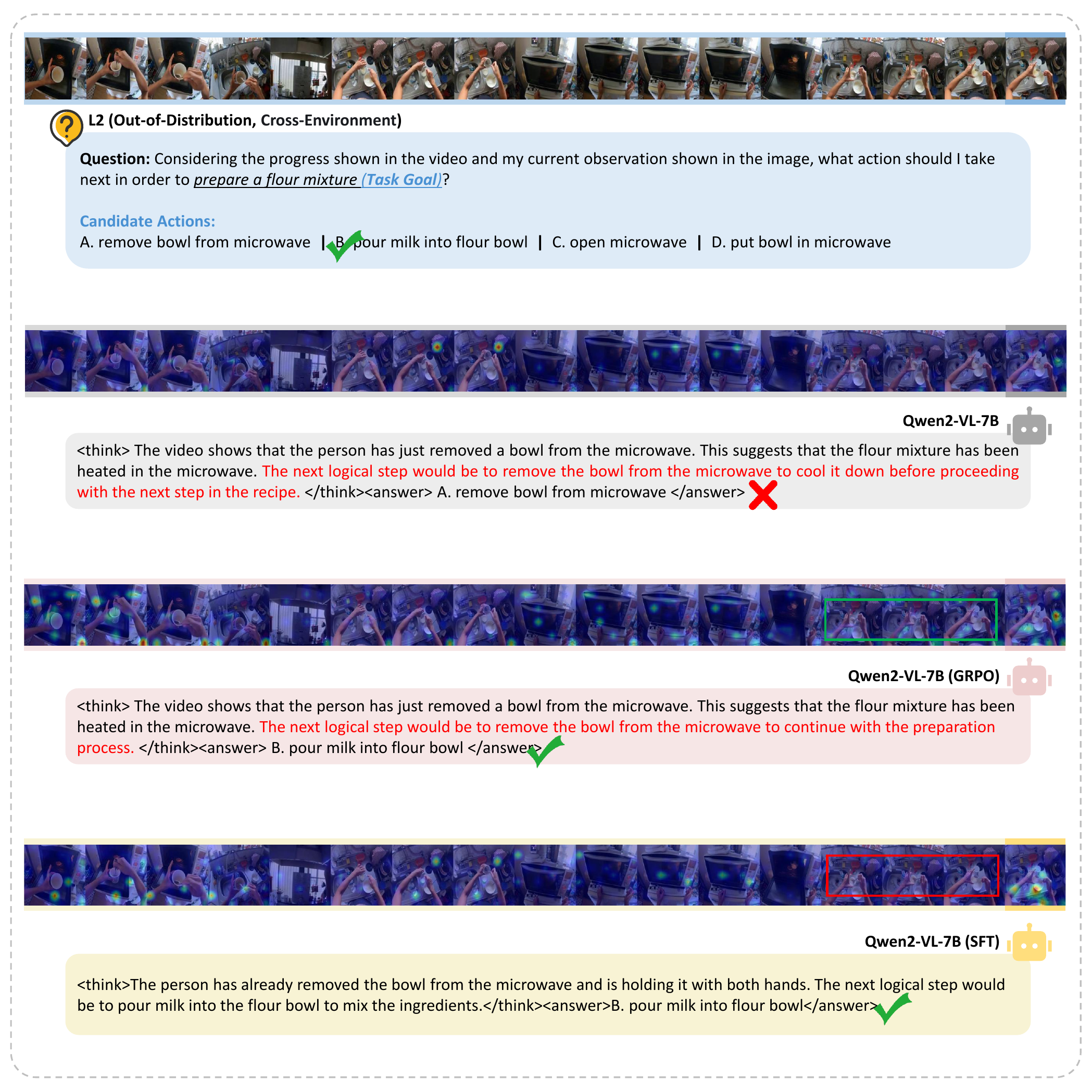}
    \caption{
    Comparison of model responses to a Level-2 (out-of-distribution, cross-environment) question from SEED-Bench-R1. The GRPO fine-tuned model \textbf{demonstrates more accurate attention} to hand movement (highlighted in the green box). Interestingly, while the GRPO fine-tuned model produces similar incorrect reasoning steps as the SFT-trained model (red text), it ultimately outputs the correct answer by disregarding the flawed semantic reasoning. This suggests that GRPO, with its outcome-supervised reward signal, primarily \textbf{enhances visual perception} but may compromise the logical coherence and semantic accuracy of the model's reasoning process.
    }
    \label{fig:case_study_L2}
\end{figure}

\subsection{Detailed Analysis}  
We analyze how RL influences the MLLM's chain-of-thought (COT) generation process and its impact on final answers, focusing on visual perception and logical inference.

\paragraph{Limited Emergent Reasoning Abilities.}
The MLLM's performance improves without significant emergent reasoning abilities. As shown in Figure~\ref{fig:reward_completion_length}, although the reward value generally climbs during RL, the completion length of the MLLM does not increase notably. This may stem from the base model's capacity.
By comparing the outputs from the base model (Qwen2-VL-7B) in Figure~\ref{fig:case_study_L1} and Figure~\ref{fig:case_study_L2}, we find that the answer accuracy does not consistently improve with longer completions pre-RL. Extended reasoning chains are not always helpful and sometimes can even introduce noise, hindering decision-making. 
Another reason may be due to the training data quality and the outcome-based reward signal. The training data provided by SEED-Bench-R1 are automatically constructed without manually verifying the uniqueness of the golden answers. Consequently, the occasional data noise makes the model confused and learn to earn outcome-based rewards by skipping the logical inference process, which may be “unreasonable” for deducing the “correct” but noisy golden answers during training.

\begin{figure}[!t]
    \centering
    \includegraphics[width=\textwidth]{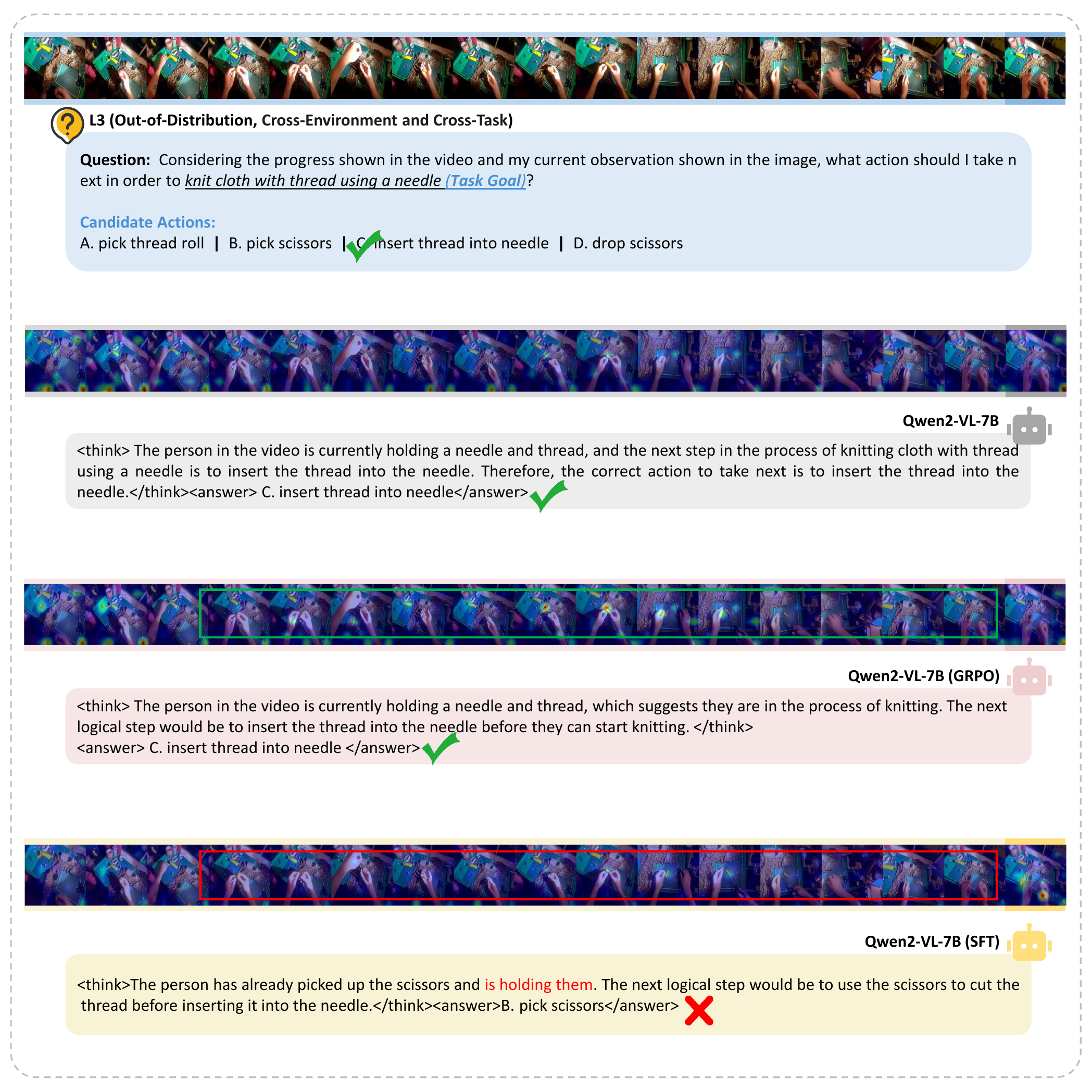}
    \caption{
    Comparison of model responses to a Level-3 (out-of-distribution, cross-environment-task) question from SEED-Bench-R1.
    The GRPO-tuned model attends effectively to the entire video, while the SFT-tuned model ignores the middle segments. SFT also favors superficial reasoning (e.g., templated phrases like ``The person has already... and is holding...''), often mismatching visual observations, which is also evidenced in Figure~\ref{fig:case_study_L1}. This suggests GRPO trains models to \textbf{search visual space adaptively}, whereas SFT \textbf{encourages memorized reasoning patterns}.
    }
    \label{fig:case_study_L3_1}
\end{figure}

\paragraph{Enhanced Visual Perception.}
Post-RL, the MLLM's COT content aids perception more than reasoning, as evidenced by the case study in Figure~\ref{fig:case_study_L1} to \ref{fig:case_study_L3_1}. Compared with the counterpart trained with SFT, the MLLM trained with RL may lack logical coherence in the generated COT content. However, by analyzing the attention map from the generated COT tokens to the visual inputs, we find that the COT tokens for the model trained with RL act more like dynamic ``queries'', attending to the visual content more comprehensively and more accurately than those for the model trained with SFT, particularly in OOD scenarios. We speculate that RL teaches the model to search more effectively in the visual space with the additional COT tokens, whereas SFT often forces memorization of reasoning patterns, leading to superficial COT content generation without sufficient attention to the visual inputs. This may explain RL's superior generalization.

\begin{figure}[!t]
    \centering
    \includegraphics[width=\textwidth]{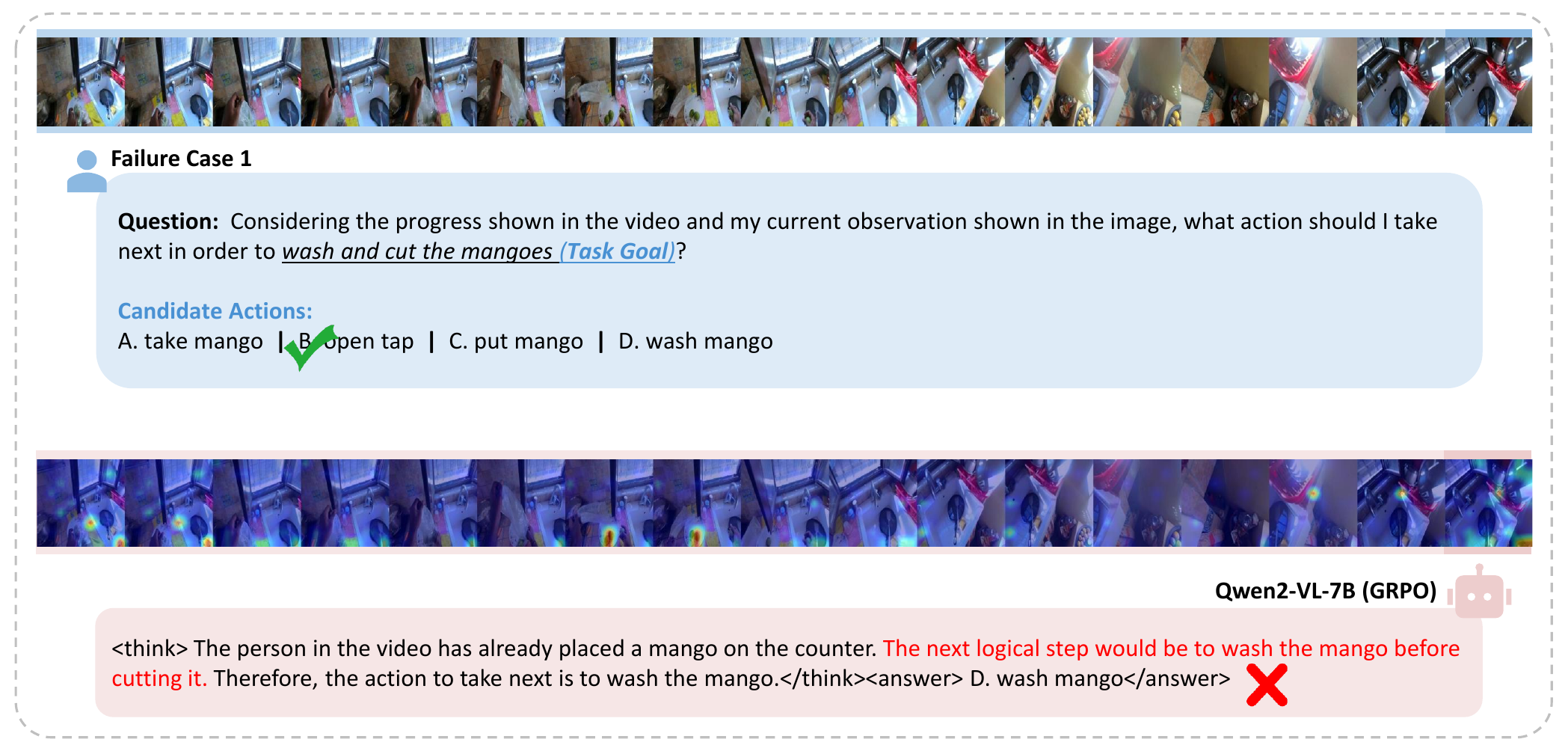}
    \caption{
    Failure Case 1. The model post-trained using GRPO-based RL \textbf{lacks commonsense reasoning}—it fails to recognize that the tap must be turned on before washing the mango.
    }
    \label{fig:failure_case_1}
\end{figure}

\begin{figure}[!t]
    \centering
    \includegraphics[width=\textwidth]{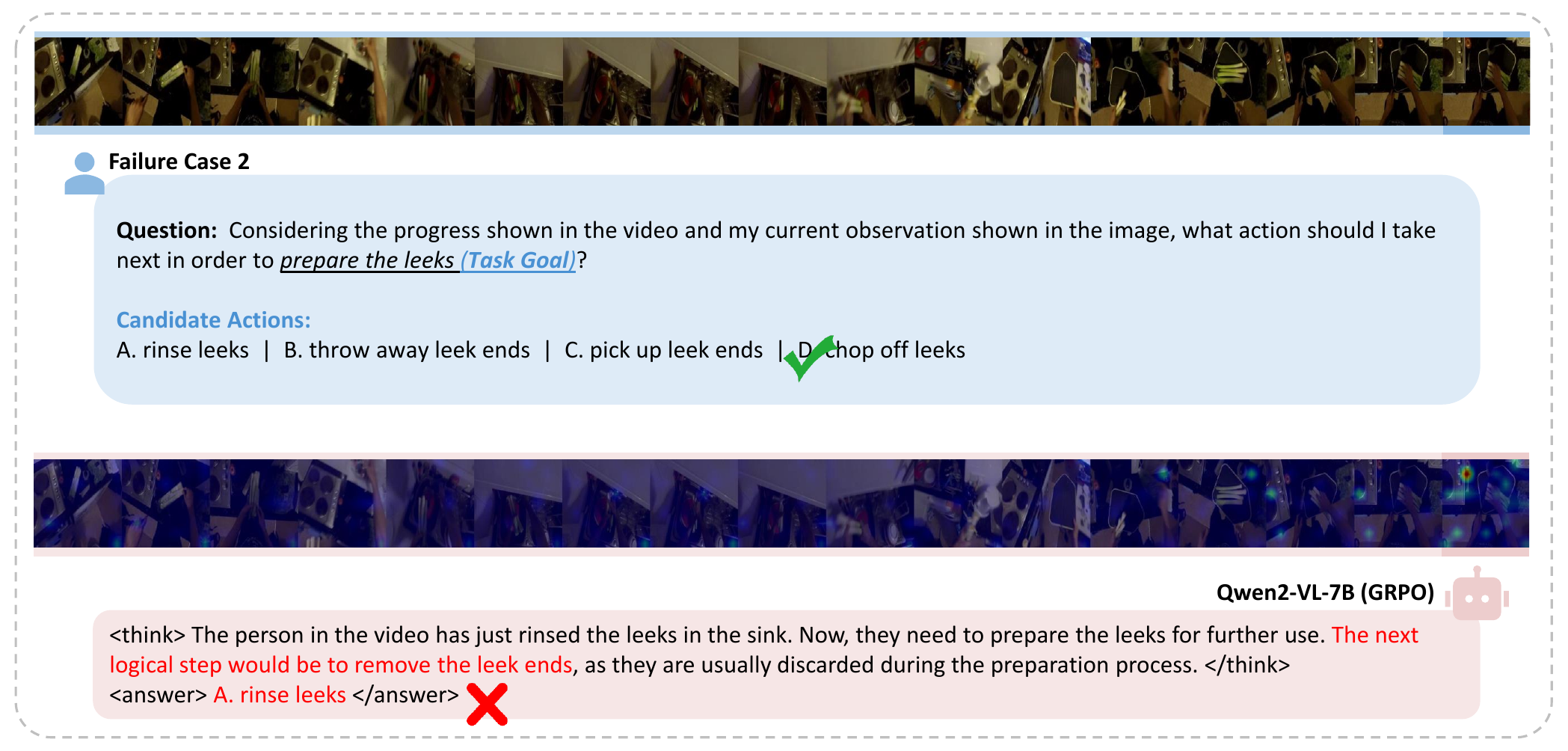}
    \caption{
    Failure Case 2. The GRPO-based RL post-trained model \textbf{overlooks key visual cues}—it fails to detect the removal of leek ends due to the limited frame sampling rate and low frame resolution. Additionally, the final answer exhibits semantic inconsistency with the reasoning steps.
    }
    \label{fig:failure_case_2}
\end{figure}

\paragraph{Limitations of Outcome Supervision RL with GRPO.}
Further, we identify the limitations of our current practice with RL through additional studies on failure cases in Figure~\ref{fig:failure_case_1} to \ref{fig:failure_case_2}. On the one hand, despite improved visual perception after RL, the model may still overlook key visual cues. 
One of the reasons may lie in constraints like frame sampling rate and image resolution, which are limited to reduce context length for training efficiency. Additionally, captions reflecting perception results typically appear only at the beginning of COT content. Ideally, the model should interleave captioning and logical inference to enable backtracking key visual areas and self-correction.  
On the other hand, the MLLM post-trained with RL using outcome supervision rewards often produces inconsistent or illogical reasoning chains. While the model may derive correct answers despite flawed COT, this undermines decision-making transparency. 
Moreover, limited reasoning ability also constrains the upper bound of model performance, since it is essential for the model to combine the internal world knowledge with visual perception results to logically infer the correct answer for more challenging questions.

\paragraph{Future Directions.}
Based on our experimental findings, we propose several promising directions for future research: 
\begin{itemize}
    \item \textbf{Pre-RL Reasoning Elicitation:} 
      Before RL, the base MLLM should exhibit a trend of generating improved answers as its chain-of-thought (COT) reasoning becomes more sophisticated. This capability enables self-evolution during RL, where higher-quality COT leads to greater rewards from outcome-based signals. Future work could investigate efficient data curation methods to collect high-quality COT demonstrations that showcase advanced reasoning skills—such as problem decomposition, reflection, and self-correction. Fine-tuning the model on such data as a cold start could streamline and enhance subsequent RL training.
    
    \item \textbf{Reward Modeling and RL Algorithms:} Regularizing the model to balance visual perception and logical inference is crucial. Process-based rewards could explicitly supervise reasoning rationality, preventing shortcuts. Additionally, we have currently only used a small proportion of training data from SEED-Bench-R1. 
    Enhancing RL algorithms' robustness against noisy reward signals is vital for scaling with larger, less clean datasets and enabling weak-to-strong alignment. 
    Moreover, improving the efficiency of algorithms to allow for a long context of visual inputs also plays an important role in large-scale training.

\end{itemize}

\section{Conclusion}

In this work, we introduce SEED-Bench-R1, a benchmark for systematically evaluating post-training methods in Multimodal Large Language Models (MLLMs), with a focus on video understanding tasks that demand both nuanced perception and robust logical reasoning. By leveraging hierarchically structured validation splits, SEED-Bench-R1 provides a rigorous testbed to assess in-distribution and out-of-distribution generalization. Our experiments with reinforcement learning (RL), specifically GRPO, demonstrated its superiority over supervised fine-tuning (SFT) in data efficiency and performance across all generalization levels, even extending to general video benchmarks like LongVideoBench. By refining reward designs, incorporating process supervision, and scaling to larger datasets, future efforts can further bridge the gap between perceptual and reasoning prowess in MLLMs, advancing their applicability in open-ended video understanding tasks.


\bibliography{iclr2025_conference}
\bibliographystyle{iclr2025_conference}


\end{document}